\pdfoutput=1

\documentclass[11pt]{article}

\usepackage[]{EMNLP2022}
\usepackage{times}
\usepackage{latexsym}

\usepackage[T1]{fontenc}

\usepackage[utf8]{inputenc}

\usepackage{microtype}

\usepackage{amsmath,amsfonts,amsthm}
\usepackage{bm}
\usepackage{booktabs}
\usepackage{multirow}
\usepackage{xcolor}
\usepackage{colortbl}
\usepackage{xspace}
\usepackage{subfig}
\usepackage{graphicx} 
\usepackage{color}
\usepackage{enumitem}
\usepackage[ruled,linesnumbered]{algorithm2e}
\usepackage{makecell}
\usepackage{float}
\usepackage{amsthm}
\theoremstyle{definition}
\newtheorem{definition}{Definition}
\usepackage{pifont}

\usepackage{enumerate}

\usepackage{paralist}

\title{MetaTKG: Learning Evolutionary Meta-Knowledge for Temporal Knowledge Graph Reasoning}

\author{\textbf{Yuwei Xia$^{1,3}$ , Mengqi Zhang$^{2,4}$ , Qiang Liu$^{2,4}$\footnotemark[1] , Shu Wu$^{2,4}$ , Xiao-Yu Zhang$^{1,3}$\footnotemark[1]} \\
  $^1$Institute of Information Engineering, Chinese Academy of Sciences \\
  $^2$Center for Research on Intelligent Perception and Computing (CRIPAC) \\
  National Laboratory of Pattern Recognition (NLPR) \\
  Institute of Automation, Chinese Academy of Sciences \\
  $^3$School of Cyber Security, University of Chinese Academy of Sciences \\
  $^4$School of Artificial Intelligence, University of Chinese Academy of Sciences \\
  \texttt{xiayuwei@iie.ac.cn, mengqi.zhang@cripac.ia.ac.cn} \\ \texttt{\{qiang.liu,shu.wu\}@nlpr.ia.ac.cn, zhangxiaoyu@iie.ac.cn} \\
 } 
  
\newcommand{\themodel}{MetaTKG\xspace}
\newcommand{\themodelAblation}{MetaTKG-G\xspace}
\begin{document}
\maketitle

\renewcommand{\thefootnote}{\fnsymbol{footnote}}
\footnotetext[1]{To whom correspondence should be addressed.} 
\renewcommand{\thefootnote}{\arabic{footnote}}

\begin{abstract}
Reasoning over Temporal Knowledge Graphs (TKGs) aims to predict future facts based on given history. One of the key challenges for prediction is to learn the evolution of facts. Most existing works focus on exploring evolutionary information in history to obtain effective temporal embeddings for entities and relations, but they ignore the variation in evolution patterns of facts, which makes them struggle to adapt to future data with different evolution patterns. Moreover, new entities continue to emerge along with the evolution of facts over time. Since existing models highly rely on historical information to learn embeddings for entities, they perform poorly on such entities with little historical information. To tackle these issues, we propose a novel Temporal Meta-learning framework for TKG reasoning, MetaTKG for brevity. Specifically, our method regards TKG prediction as many temporal meta-tasks, and utilizes the designed Temporal Meta-learner to learn evolutionary meta-knowledge from these meta-tasks. The proposed method aims to guide the backbones to learn to adapt quickly to future data and deal with entities with little historical information by the learned meta-knowledge. Specially, in temporal meta-learner, we design a Gating Integration module to adaptively establish temporal correlations between meta-tasks. Extensive experiments on four widely-used datasets and three backbones demonstrate that our method can greatly improve the performance.
\end{abstract}

\section{Introduction}
Temporal Knowledge Graphs (TKGs) \cite{DVN/28075_2015} are of great practical values as an effective way to represent the real-world time-evolving facts. In TKGs, the expression of a fact is extended from a triple to a quadruple, such as $\left ( Obama, run\; for, president, 2012 \right )$. Due to the great practical values of TKGs, related research on TKG reasoning has continued to emerge in recent years. In this paper, we mainly focus on predicting future facts over TKGs, which is beneficial for a wide range of practical scenarios such as health care and financial analysis. Specifically, given a TKG that contains historical facts from $t_{0}$ to $t_{n}$, we aim to predict new facts at timestamp $t$ with $t > t_{n}$, such as $\left ( Obama, visit, ?, 2014 \right )$.  

The vast majority of existing solutions on TKG reasoning \cite{DBLP:conf/icml/TrivediDWS17, DBLP:conf/iclr/TrivediFBZ19, 2020Recurrent, DBLP:conf/sigir/LiJLGGSWC21} mainly focus on exploring evolutionary information in history to obtain effective temporal embeddings for entities and relations. However, it is far from sufficient to only consider the temporal information for the complex evolution process of facts. 

\begin{figure}[t]
	\centering
	\subfloat[]{\includegraphics[scale=0.55]{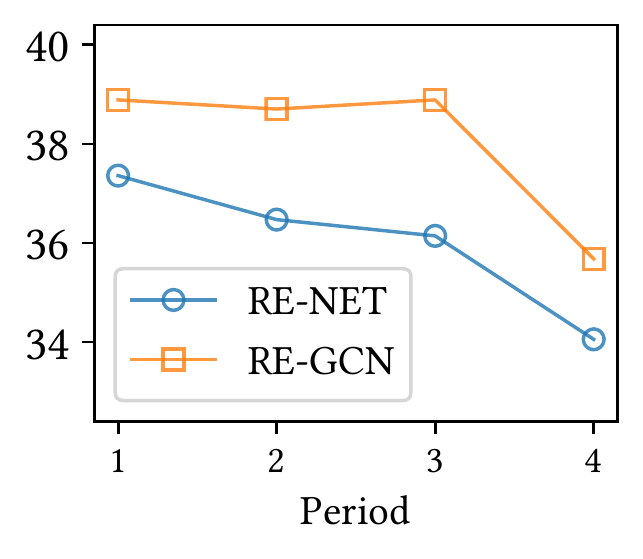}\label{mov1}
	}
	\subfloat[]{\includegraphics[scale=0.55]{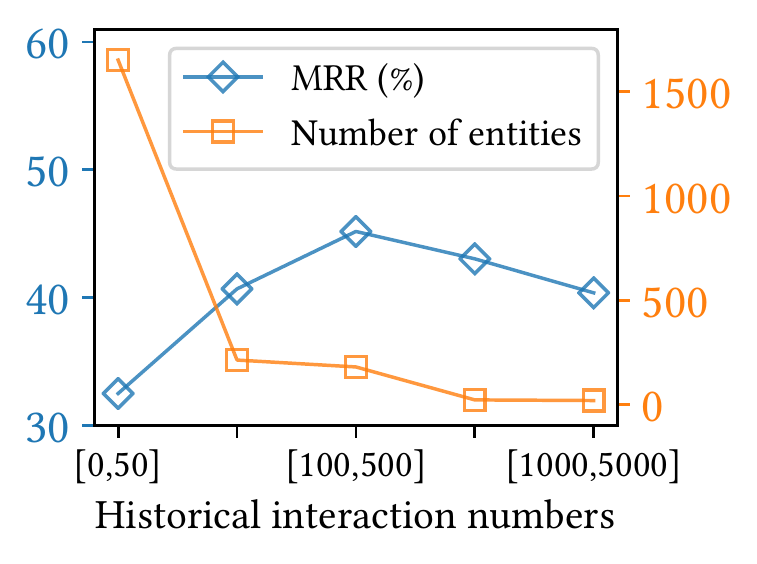}\label{mov2}
	}
	\caption{
	(a) The y-axis is the performance (MRR (\%)) of existing TKG models RE-NET and RE-GCN on predicting facts in different time periods. (b) The left y-axis is the performance of RE-NET on predicting entities with different historical interaction numbers. And the right y-axis is the number of entities in each group. The results of both (a) and (b) are obtained based on the widely used TKG dataset ICEWS14.
	}
\end{figure}

Specifically, due to the diversity of facts, they exhibit various evolution patterns in the evolution process. This brings up the issue that most models are hard to adapt to future data with different evolution patterns. For example, models trained with facts before COVID-19 are hard to adapt to facts that happen under the circumstance of COVID-19, because the evolution pattern of facts changes with the outbreak of COVID-19. Thus, it is important to learn variations in evolution patterns so as to guide models quickly adapt to future data with diverse evolution patterns. However, existing TKG reasoning models ignore the learning for evolution patterns. From Figure \ref{mov1}, we can see the performance of existing TKG models drops along time, because models trained on historical data learn the old pattern which is not applicable to future data \cite{DBLP:conf/cikm/YouZDFH21}.

Moreover, as facts evolve constantly over time, there will be new entities continuing to emerge during the evolution of facts. Most of these new entities have little historical information and thus are hard to learn. As shown in Figure \ref{mov2}, we can see that an extremely great number of future entities appear less than 50 times during evolution, and the performance of existing methods on such entities is much worse than others. Hence, improving the performance on entities with little historical information is necessary for prediction. However, most existing methods fail to address the issue, because they highly rely on sufficient historical information to obtain effective temporal embeddings of entities.

To deal with the aforementioned challenges, we propose a novel Temporal meta-learning framework for TKG reasoning, MetaTKG for brevity. MetaTKG is plug-and-play and can be easily applied to most existing backbones for TKG prediction. Specifically, MetaTKG regards TKG prediction as many temporal meta-tasks for training, and utilizes a Temporal Meta-Learner to learn the evolutionary meta-knowledge from these meta-tasks. Specially, each task consists of two KGs with adjacent time stamps in TKGs. In this way, the temporal meta-learner can learn the variation in evolution patterns of two temporally adjacent KGs, which can guide the backbones to learn to adapt quickly to future data with different evolution patterns. Besides, from the learning process of each task, the backbones can derive the meta-knowledge learned from entities with sufficient history to study how to obtain effective embeddings for entities with little historical information. Moreover, we specially design a Gating Integration module in temporal meta-learner according to the characteristics of TKGs, to adaptively establish the temporal correlations between tasks during the learning process. 

To summarize, the major contributions can be listed as follows:
\begin{itemize}
\item We illustrate and analyze the critical importance of learning the variations in evolution patterns of facts and new entities with little historical information on TKG reasoning.
\item We propose a novel meta-learning framework to learn the evolutionary meta-knowledge for TKG prediction, which can be easily plugged into most existing TKG backbones.
\item We conduct extensive experiments on four commonly-used TKG benchmarks and three backbones, which demonstrate that \themodel can greatly improve the performance.
\end{itemize}

\section{Related Work}
There are mainly two settings in TKG reasoning: interpolation and extrapolation. In this paper, we focus on the latter setting, which aims to predict future facts based on the given history. And we briefly introduce the applications of meta-learning in KGs and TKGs in this section.

\textbf{TKG Reasoning with extrapolation setting.} 
Several early attempts such as GHNN \cite{DBLP:conf/akbc/HanM0GT20}, Know-Evolve \cite{DBLP:conf/icml/TrivediDWS17} and DyRep \cite{DBLP:conf/iclr/TrivediFBZ19} build a temporal point process to obtain continuous-time dynamics for TKG reasoning, but they merely focus on continuous-time TKGs. Recently, RE-NET \cite{2020Recurrent} models TKGs as sequences, utilizing R-GCNs and RNNs to capture structural and global temporal information, respectively. Following RE-NET, RE-GCN \cite{DBLP:conf/sigir/LiJLGGSWC21} captures more complex structural dependencies of entities and relations, and utilizes static information to enrich embeddings of entities. Besides, CyGNet \cite{DBLP:conf/aaai/ZhuCFCZ21} proposes a copy-generation mechanism, which utilizes recurrence patterns in history to obtain useful historical information. TANGO \cite{han-etal-2021-learning-neural} applies the Neural Ordinary Differential Equations (NODEs) on TKGs to capture continuous temporal information. And CEN \cite{DBLP:conf/acl/LiGJPL000GC22} considers the length diversity of history and explores the most optimal history length of each dataset for prediction. Moreover, xERTE \cite{DBLP:conf/iclr/HanCMT21} designs an inference graph to select important entities for prediction, which visualizes the reasoning process. And TITer \cite{sun-etal-2021-timetraveler} presents a path-search model based on Reinforcement Learning for TKG prediction. However, all of these models are unable to well adapt to new data with diverse evolution patterns, and fail to learn entities with little historical information. 

\textbf{Meta-Learning in KGs.} Meta-learning is regarded as “learning to learn" \cite{DBLP:journals/air/VilaltaD02}, which aims to transfer the meta-knowledge so as to make models rapidly adapt to new tasks with some examples. Meta-learning has been wildly used in various fields. In KGs, meta-learning has also been verified its effectiveness. GMatching \cite{DBLP:conf/emnlp/XiongYCGW18} proposes the problem of few-shot relation in KGs, and applies a metric-based meta-learning to solve the issue by transferring the meta-knowledge from the background information to few-shot relation. Afterward, MetaR \cite{DBLP:conf/emnlp/ChenZZCC19} presents a solution for few-shot relation in KGs following the meta-learning framework, which is independent of the background knowledge compared to GMatching. Recently, some works which follow GMatching to solve the problem of few-shot relation in KGs have emerged, such as FSRL \cite{DBLP:conf/aaai/ZhangYHJLC20}, FAAN \cite{DBLP:conf/aaai/ZhangYHJLC20} and GANA \cite{DBLP:conf/sigir/NiuLTGDLWSHS21}. These works design different neighbor aggregation modules to make improvements on representation learning. However, all of these works are difficult to be adopted for TKGs. OAT \cite{DBLP:conf/akbc/MirtaheriR0MG21} is the first attempt to utilize meta-learning to solve the problem of one-shot relation on TKGs, which performs on newly constructed one-shot TKG dataset. OAT generates temporal embeddings of entities and relations with a transformer-based encoder \cite{DBLP:conf/nips/VaswaniSPUJGKP17}. And the utilization in OAT of meta-learning is similar to KGs, which divides tasks by relations. 

Different from these meta-learning works mentioned above, our proposed method focuses on entities with little historical information in TKGs. Moreover, our model does not need to perform on special datasets, which is of more practical values compared to the models that only center on few-shot relations. Besides, though OAT utilizes the meta-learning method, the way it divides tasks still results in its inability to learn the variation in evolution patterns.

\begin{figure*}
  \centering
  \includegraphics[width=\linewidth]{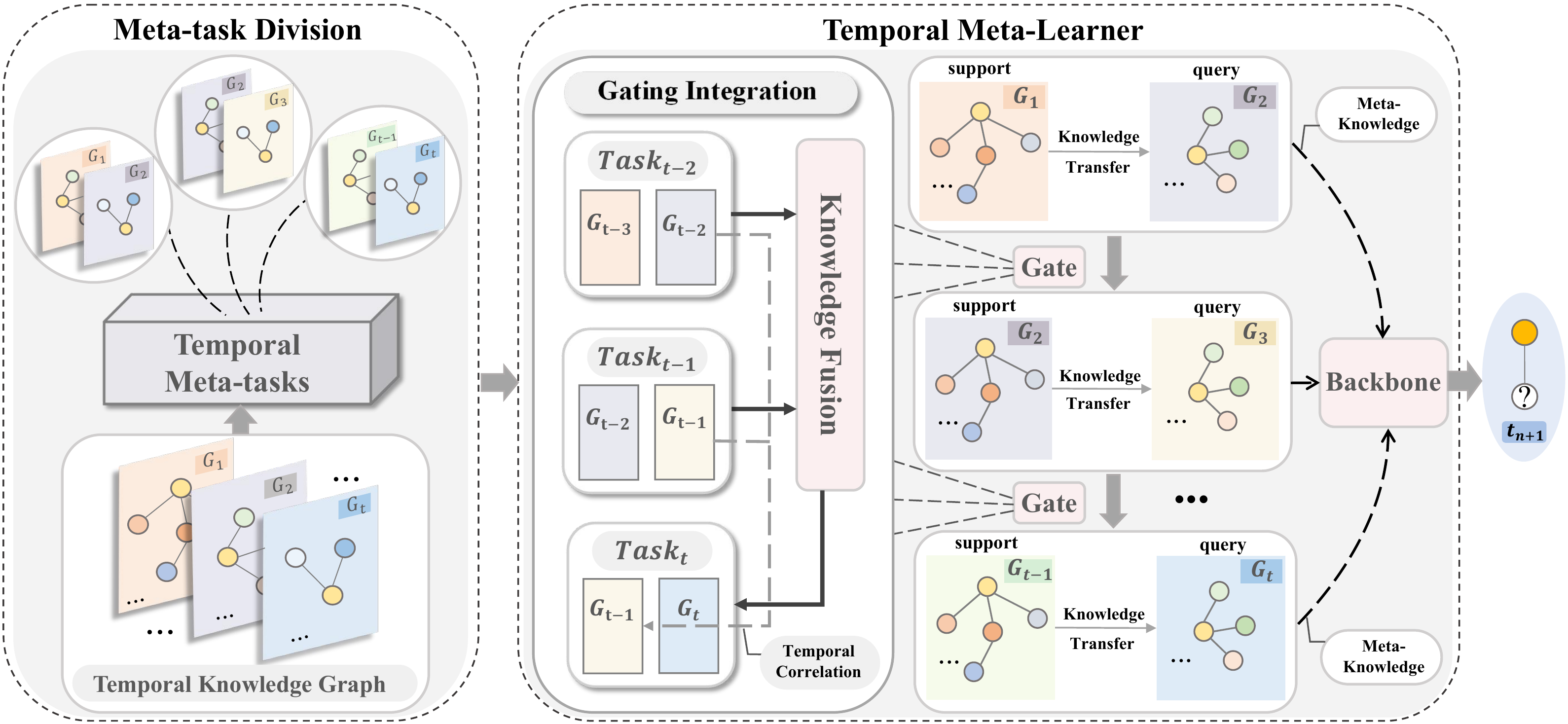}
  \caption{
  An illustration for \themodel. The TKG is firstly divided into many \emph{temporal meta-tasks} (\S \ref{temporal task}) for training. Then the \emph{Temporal meta-learner} (\S \ref{Temporal Meta-learner}) learns the evolutionary meta-knowledge from these tasks. Specially, according to the temporal characteristics of TKGs, we design a \emph{Gating Integration} (\S \ref{gate}) module to establish the temporal correlations among these tasks. During the learning process of each task, the knowledge gained from former ones will be fused and then transferred to the latter task. Finally, the obtained meta-knowledge from different tasks guides the backbone to learn to quickly adapt to new data for prediction over TKGs.
 }
 \label{framework}
\end{figure*}

\section{Preliminaries}
\label{definition}
In this section, we mainly formulate TKGs and the problem of TKG reasoning, then give brief notations used in this work.
\begin{definition}[Temporal Knowledge Graph]
Let $\mathcal{E}$ and $\mathcal{R}$ represent a set of entities and relations. A Temporal Knowledge Graph (TKG) $\mathcal{G}$ can be defined as a temporal sequence consisting of KGs with different timestamps, i.e., $\mathcal{G} = \{\mathcal{G}_1, \mathcal{G}_2, \cdots, \mathcal{G}_n\}$. Each $\mathcal{G}_t \in \mathcal{G}$ contains facts that occur at time $t$. And a fact is described as a quadruple $(e_s,r,e_o, t)$, in which $e_s, e_o \in \mathcal{E}$ and $r \in \mathcal{R}$.
\end{definition}

\begin{definition}[TKG Reasoning]
The task of TKG Reasoning with extrapolation setting can be categorized into \emph{entity prediction} and \emph{relation prediction}. The \emph{entity prediction} task aims to predict the missing object entity of $(e_s, r, ?, t+1)$ or the missing subject entity of $(?, r, e_o, t+1)$ given historical KG sequence $\{\mathcal{G}_1, \mathcal{G}_2, \cdots, \mathcal{G}_t\}$. Similarly, the \emph{relation prediction} task aims to predict the missing relation of $(e_s, ?, e_o, t+1)$. In this paper, we mainly evaluate our models on the entity prediction task.
\end{definition}

\begin{definition}[Backbones for TKG Reasoning]
We denote TKG reasoning backbones as a parametrized function $f_{\theta}$ with parameters $\theta$.
\end{definition}

\begin{definition}[Meta-task]
In the traditional setting of meta-learning \cite{DBLP:conf/icml/FinnAL17}, the training process is based on a set of meta-tasks. We denote a meta-task as $\mathcal{T}_t$. Each meta-task consists of a support set $\mathcal{T}_t^s$ and a query set $\mathcal{T}_t^q$, which can be denoted as $\mathcal{T}_t=\{\mathcal{T}_t^s, \mathcal{T}_t^q\}$. During the training process of each task $\mathcal{T}_t$, the backbones are first trained on $\mathcal{T}_t^s$. Then the backbones are trained on $\mathcal{T}_t^q$ with the feedback from the loss of $\mathcal{T}_t^s$. 
\end{definition}

\section{Proposed Approach}
In this section, we present the proposed \themodel in detail. The framework of our model is shown in Figure \ref{framework}. In \themodel, TKG prediction is regarded as many \emph{temporal meta-tasks} for training. The \emph{temporal meta-learner} in our model is built to learn evolutionary meta-knowledge from these meta-tasks. We aim to guide the backbones to learn to adapt quickly to future data and learn to deal with entities with little historical information by the learned meta-knowledge. Moreover, in the \emph{temporal meta-learner}, the \emph{Gating Integration} module shown in the middle part of Figure \ref{framework} is specially designed to adaptively capture the temporal correlation between tasks.

\subsection{Temporal Meta-tasks}
\label{temporal task}
In TKGs, the temporal correlation between $\mathcal{G}_{t-1}$ and $\mathcal{G}_{t}$ implies the evolutionary changes. According to this, we can learn the variation in evolution patterns by learning the evolutionary information in $\mathcal{G}_{t-1}$ and $\mathcal{G}_{t}$. Thus, in our model, we regard TKG prediction as many temporal meta-tasks $\mathcal{T}_t$ for training. Each task is designed to consist of two KGs in $\mathcal{G}$ with adjacent time stamps, which is more applicable to the TKG scenario. Formally, a temporal meta-task $\mathcal{T}_t$ can be denoted as:
\begin{equation}
    \mathcal{T}_t = \{\mathcal{G}_{t-1}, \mathcal{G}_{t}\},
\end{equation}
where $\mathcal{G}_{t-1}$ is the support set $\mathcal{T}_t^s$ and $\mathcal{G}_{t}$ is the query set $\mathcal{T}_t^q$. That is, our training data can be described as $\mathbb{T}_{train} = \{\mathcal{T}_t\}_{t=2}^{k}$, where each task $\mathcal{T}_t$ corresponds to an individual entity prediction task. Similarly, the validation data and testing data are also composed of temporal meta-tasks, which can be denoted as $\mathbb{T}_{valid}= \{\mathcal{T}_t\}_{t=k+1}^{m}$ and $\mathbb{T}_{test}= \{\mathcal{T}_t\}_{t=m+1}^{n}$, respectively.

\subsection{Temporal Meta-learner}
\label{Temporal Meta-learner}
To learn the evolutionary meta-knowledge from the temporal meta-tasks, inspired by the learning ability of meta-learning in the time series scene \cite{DBLP:conf/cikm/YouZDFH21, DBLP:conf/wsdm/XieWWLZ0L22}, we design a novel temporal meta-learner according to the characteristics of TKGs. The goal of the temporal meta-learner is to guide the backbones quickly adapt to future data by the learned meta-knowledge.

In specific, during the learning process of each $\mathcal{T}_t$, we first train the backbones on the support set $\mathcal{T}_t^s$. And the updated parameter is computed by using one gradient update. Formally, this process can be defined as follows,
\begin{equation}
\label{inner}
    {\theta}'_t = \theta_t^s - \alpha \nabla_{\theta_t^s} \mathcal{L}_{\mathcal{T}_t^s}\left(f_{\theta_t^s}\right),
\end{equation}
where ${\theta}'_t$ is the updated parameter on the support set $\mathcal{T}_t^s$, and $\theta_t^s$ represents the initial parameter for training the backbones on $\mathcal{T}_t^s$. $\alpha$ is a hyper-parameter, which is for controlling the step size.

After obtaining ${\theta}'_t$ updated on $\mathcal{T}_t^s$ and the feedback from the loss of $\mathcal{T}_t^s$, we train the backbones on the query set $\mathcal{T}_t^q$ by
\begin{equation}
\label{outer}
    {\theta}_t = \theta_t^q - \beta \nabla_{\theta_t^q} \mathcal{L}_{\mathcal{T}_t^q}\left(f_{{\theta}^{'}_t}\right),
\end{equation}
where ${\theta}_t$ is the updated parameter on the query set of $\mathcal{T}_t$. And $\theta_t^q$ represents the initial parameter for training the backbones on $\mathcal{T}_t^q$. $\beta$ is a hyper-parameter for controlling the step size. It is to be noted that we learn one meta-task at a time, the initial parameter for learning each $\mathcal{T}_t$ is the updated parameter by previously learned meta-tasks, which will be further described in detail in \S \ref{gate}. 

By utilizing such training strategy, the evolutionary meta-knowledge learned from each $\mathcal{T}_t$ can guide the backbones to gradually learn to face new data (the query set $\mathcal{T}_t^q$) with different evolution patterns from the old data (the support set $\mathcal{T}_t^s$). In this way, by continuously learning these tasks one by one, the backbones can learn to quickly adapt to new data by the accumulated meta-knowledge from different tasks. Meanwhile, the meta-knowledge also guides the backbones to learn entities with little historical information with the learning experience from other entities. 

\subsubsection{Gating Integration}
\label{gate}
Due to the temporal characteristics of TKGs, meta-tasks in TKG scenario are temporally correlated. That is, the meta-knowledge learned from former meta-tasks is helpful for learning the next one. Thus, in the learning process of meta-tasks, the temporal correlation between them is necessary to be considered. Since such correlation is actually generated by the temporal correlation between KGs with adjacent time stamps in TKGs, the key to establishing the correlation between meta-tasks is to associate temporally adjacent KGs in different tasks.

Considering the importance of establishing the temporal correlations, we specially design a gating integration module to effectively build up the temporal correlations between tasks. 

Specifically, for the support set of each task, we fuse the updated parameter vector by task $\mathcal{T}_{t-1}$ and $\mathcal{T}_{t-2}$, taking the fused one as the initial parameter of task $\mathcal{T}_t$ for learning. Formally, the initial parameter $\theta_t^s$ in Eq (\ref{inner}) can be calculated as,
\begin{equation}
\label{gate-supp}
    \theta_t^s = \sigma\left(\operatorname{g}_s\right) \odot \theta_{t-1} + \left(1-\sigma\left(\operatorname{g}_s\right)\right) \odot \theta_{t-2},
\end{equation}
where $\operatorname{g}_s$ is a learn-able gate vector to balance the information of $\theta_{t-1}$ and $\theta_{t-2}$. $\sigma(\cdot)$ is the Sigmoid function to project the value of each element into $[0,1]$, and $\odot$ denotes element-wise multiplication. 
It is important to note that the parameter $\operatorname{g}_s$ is updated with the loss of the support set $\mathcal{T}_t^s$ in Eq (\ref{inner}). Formally, 
\begin{equation}
\label{g-update}
    \operatorname{g}_s \leftarrow \operatorname{g}_s - \alpha \nabla_{\operatorname{g}_s} \mathcal{L}_{\mathcal{T}_t^s}\left(f_{{\theta}_t^s}\right).
\end{equation}

By initializing $\theta_t^s$ with the gating module, we can achieve building up temporal correlations for the support set in each task. As shown in the gating integration module of Figure \ref{framework}, $\theta_t^s$ operated on $\mathcal{G}_{t-1}$ (the support set of $\mathcal{T}_t$) can contain knowledge in $\theta_{t-2}$ learned from $\mathcal{G}_{t-2}$ (the query set of $\mathcal{T}_{t-2}$). Finally, the temporal correlations can be established by temporally associating $\mathcal{G}_{t-1}$ and $\mathcal{G}_{t-2}$ in different tasks. More discussions on temporal correlation for $\mathcal{T}_t^s$ can be seen in Appendix \ref{temproal connetion}.

Different from the support set, we simply take the updated parameter $\theta_{t-1}$ by $\mathcal{T}_{t-1}$ as the initial parameter $\theta_t^q$ in Eq (\ref{outer}) for learning $\mathcal{T}_t^q$: 
\begin{equation}
\label{gate-query}
    \theta_t^q = \theta_{t-1}.
\end{equation}
Noted that the query set $\mathcal{T}_t^q = \mathcal{G}_{t}$ of $\mathcal{T}_t$ is temporally adjacent with the query set $\mathcal{T}_{t-1}^q = \mathcal{G}_{t-1}$ of task $\mathcal{T}_{t-1}$. Thus, by simply initializing $\theta_t^q$ with $\theta_{t-1}$ which contains knowledge learned from $\mathcal{G}_{t-1}$ (the query set of $\mathcal{T}_{t-1}$), we can establish temporal correlation without gating for $\mathcal{T}_t^q$ by associating $\mathcal{G}_{t}$ and $\mathcal{G}_{t-1}$ from different tasks.

\subsubsection{Component-specific gating parameter}
In particular, the gating parameter $\operatorname{g}_s$ in the gating integration module is designed to be component-specific, because we consider that parameters in different components should be updated with different frequencies. For example, the parameter of entity embeddings is updated only when the entity appears, the parameter of relation embeddings is updated more frequently, and other parameters in models need to be updated in every meta-task. Thus, we assign different $\operatorname{g}_s$ values to entity embedding, relation embedding, and other parameters in the model, respectively.

\subsection{Model Evaluation}
After training the backbones on all the tasks in $\mathcal{T}_{train}$, we can obtain the final updated parameter $\theta_{train}$ by $\mathcal{T}_{train}$ for testing. The process of testing is similar to training. Specially, to ensure the consistency in time series, we update the parameter $\theta_{train}$ on $\mathcal{T}_{valid}$ before testing. Moreover, to enhance the ability of fast adaptions for the backbones plugged with our method, we conduct multi-step gradient updates in the testing phase \cite{DBLP:conf/icml/FinnAL17}. Algorithm \ref{alg} provides the pseudo-code of the overall framework.

\SetKwComment{Comment}{/* }{ */}
\SetKwInput{KwData}{Input}
\SetKwInput{KwResult}{Output}
\SetNoFillComment
\begin{algorithm}[t]
\caption{Training procedure}\label{alg}
\KwData{$\mathcal{T}_{train}$: $\{\mathcal{T}_t\}_{t=2}^{k}$;Initial parameters $\theta$, $\operatorname{g}_s$.}
\KwResult{The trained parameters $\theta_{train}$, $\operatorname{g}_s$.}
Initialize ${\theta}$ and $\operatorname{g}_s$ randomly;

\While{\textnormal{not converged}}{
		
		\For{task $\mathcal{T}_t$ in $\mathcal{T}_{train}$}{
			Initialize $\theta_t^s$ for learning the support set $\mathcal{T}_t^s$ by Eq (\ref{gate-supp});
			
			Evaluate $\mathcal{L}_{\mathcal{T}_t^s}\left(f_{{\theta_t^s}}\right)$ on $\mathcal{T}_t^s$;
			
			Update $\theta_t^s$ and $\operatorname{g}_s$ by Eq (\ref{inner}) and Eq (\ref{g-update}) to obtain the output ${\theta}_t^{\prime}$;
			
			Initialize $\theta_t^q$ by Eq (\ref{gate-query}) before learning $\mathcal{T}_t^q$;
			
			Evaluate $\mathcal{L}_{\mathcal{T}_t^q} \left(f_{{\theta}_t^{\prime}}\right)$ by ${\theta}_t^{\prime}$ on $\mathcal{T}_t^q$;
			
			Update $\theta_t^q$ by Eq (\ref{outer}) to obtain the output $\theta_t$;

		}
	}
	Obtain the final updated parameter on $\mathcal{T}_{train}$ as $\theta_{train}$ for testing;
	
	return $\theta_{train}$
\end{algorithm}

\section{Experiment}
\label{experiment}
In this section, we conduct experiments to evaluate \themodel on four typical datasets of temporal knowledge graph and three backbones for TKG prediction. The implementation details can be seen in Appendix \ref{implementation details}.
Then we answer the following questions through experimental results and analyses.  
\begin{itemize}
\item {\bf Q1}: How does the proposed \themodel perform when plugged into existing TKG reasoning models for the entity prediction task?
\item {\bf Q2}: How effective is \themodel on learning the variations in evolution patterns?
\item {\bf Q3}: How effective is \themodel on learning entities with little historical information?
\item {\bf Q4}: How does the Gating Integration module affect the performance of \themodel?
\item {\bf Q5}: How does the number of gradient update steps affect the performance of \themodel?
\end{itemize}

\subsection{Experimental Settings}
\label{Experimental Settings}

\subsubsection{Datasets}
We utilize WIKI \cite{DBLP:conf/www/LeblayC18} and three datasets from the Integrated Crisis Early Warning System \cite{DVN/28075_2015}: ICEWS14 \cite{A2018Learning}, ICEWS18 \cite{2020Recurrent} and ICEWS05-15 \cite{A2018Learning} to evaluate the effectiveness of our model. Each of the ICEWS dataset is composed of various political facts with time annotations such as (\emph{Barack Obama}, \emph{visit}, \emph{Malaysia}, $2014/02/19$). ICEWS14, ICEWS18, and ICEWS05-15 record the facts in 2014, 2018, and the facts from 2005 to 2015, respectively. Following \cite{DBLP:conf/sigir/LiJLGGSWC21}, We divide ICEWS14, ICEWS18, ICEWS05-15, and WIKI into training, validation, and testing sets with a proportion of 80\%, 10\%, and 10\% by timestamps. The details of the four datasets are presented in Table \ref{tab:dataset}, where the time gap represents time granularity between two temporally adjacent facts.

\begin{table}[t]
	\centering
	\caption{The statistics of the datasets.}
	\resizebox{\linewidth}{!}
	{\begin{tabular}{ccccc}
			\toprule
			 Datasets &  ICEWS14 &  ICEWS05-15 & ICEWS18 &  WIKI \\
			\midrule
			 $\#$ $\mathcal{E}$ & 6,869     & 10,094 & 23,033&12,554 \\
			$\#$ $\mathcal{R}$ & 230    & 251 & 256   &  24 \\
			$\#$ Train & 74,845 & 368,868& 373,018 &539,286  \\
			 $\#$ Valid & 8,514   &46,302  &45,995 &67,538 \\   
			$\#$ Test  &7,371  &46,159  &49,545   &63,110 \\
			Time gap  &24 hours  &24 hours &24 hours &1 year \\
			\bottomrule
	\end{tabular}}
	\label{tab:dataset}
\end{table}

\begin{table*}[!ht]
	\centering
	\caption{Performance comparison of \themodel when plugged into different backbones on four datasets in terms of MRR (\%), Hit@1 (\%), and Hit@10 (\%) (all results are under raw metrics). The highest performance is highlighted in bold. The backbones with * represent the fine-tuned one. $\Delta$\emph{Improve} and $\Delta$\emph{Improve}* indicate the relative improvements over the original backbones and fine-tuned backbones in percentage, respectively.}
	\resizebox{\textwidth}{!}{
		\begin{tabular}{ccccccccccccc}
			\toprule
			\multirow{2.5}[0]{*}{Model}&\multicolumn{3}{c}{ICEWS14} & \multicolumn{3}{c}{ICEWS18}  & \multicolumn{3}{c}{ICEWS05-15} & \multicolumn{3}{c}{WIKI} \\
			\cmidrule(lr){2-4} \cmidrule(lr){5-7} \cmidrule(lr){8-10} \cmidrule(lr){11-13}
			&MRR &  Hit@1 & Hit@10 &MRR &Hit@1  & Hit@10 & MRR &Hit@1 & Hit@10 &MRR &Hit@1  & Hit@10 \\
			\midrule 
	  	    RE-NET &38.75 &28.96 &57.61 &28.72 &18.84 &48.18 &44.05 &33.22 &65.02 &50.81 &40.52 &68.59\\
			RE-NET* &39.18 &29.43 &57.94 &28.76 &18.79 &48.49 &47.44 &36.13 &68.36 &52.17 &41.37 &70.32\\
			RE-NET + \themodel &\textbf{40.87} &\textbf{30.84} &\textbf{60.11} &\textbf{30.19} &\textbf{19.85} &\textbf{50.62} &\textbf{49.15} &\textbf{37.56} &\textbf{70.75} &\textbf{53.28} &\textbf{42.17} &\textbf{71.96} \\
			\rowcolor{gray!20}$\Delta$\emph{Improve}  &5.47\% &6.50\%	&4.34\%	&5.11\%	&5.35\%	&5.07\%	&11.57\%	&13.07\%	&8.82\%	&4.86\%	&4.06\%	&4.92\% \\
			\rowcolor{gray!20}$\Delta$\emph{Improve}* &4.30\% &4.78\% &3.75\% &4.97\% &5.62\% &4.39\%	&3.61\%	&3.97\%	&3.50\%	&2.13\%	&1.93\%	&2.33\% \\
			\midrule 
			RE-GCN &41.33	&30.61	&62.31	&31.08	&20.44	&52.06	&46.89	&35.5	&68.4	&51.53	&41.1 &69.56\\
			RE-GCN* &41.72	&31.17	&62.41	&31.02	&20.24	&52.15	&48.98	&37.2	&70.91	&52.91	&41.96	&71.32\\
			RE-GCN + \themodel &\textbf{42.76}	&\textbf{32.08}	&\textbf{63.32}	&\textbf{31.6}	&\textbf{20.85}	&\textbf{52.79}	&\textbf{50.00}	&\textbf{38.17}	&\textbf{71.85}	&\textbf{53.54}	&\textbf{42.24}	&\textbf{72.51}\\
			\rowcolor{gray!20}$\Delta$\emph{Improve} &3.46\%	&4.80\%	&1.62\%	&1.67\%	&2.01\%	&1.40\%	&6.63\%	&7.52\%	&5.04\%	&3.90\%	&2.77\%	&4.24\% \\
			\rowcolor{gray!20}$\Delta$\emph{Improve}* &2.49\%	&2.92\%	&2.54\%	&1.87\%	&3.01\%	&1.77\%	&2.08\%	&2.61\%	&2.03\%	&1.19\%	&0.67\%	&1.67\% \\
			\midrule 
			CEN &40.65	&30.36	&60.2	&28.5	&18.6	&48.01	&45.27	&34.18	&66.46	&49.31	&38.96	&67.62\\	
			CEN* &40.38	&30.25	&59.73	&28.64	&18.78	&48.31	&48.28	&36.67	&69.89 &52.69	&41.53 &71.62\\
			CEN + \themodel &\textbf{41.64}	&\textbf{31.55}	&\textbf{60.58}	&\textbf{29.7}	&\textbf{19.93}	&48.06	&\textbf{48.77}	&\textbf{37.14}	&\textbf{70.33}	&\textbf{53.42}	&\textbf{42.31} &\textbf{71.99} \\
			\rowcolor{gray!20}$\Delta$\emph{Improve} &2.44\%	&3.92\%	&0.63\%	&4.21\%	&7.15\%	&0.10\%	&7.73\%	&8.66\%	&5.82\%	&8.34\%	&8.60\%	&6.46\% \\
			\rowcolor{gray!20}$\Delta$\emph{Improve}* &3.12\%	&4.30\%	&1.42\%	&3.70\%	&6.12\%	&-0.52\% &1.01\%	&1.28\%	&0.63\%	&1.39\%	&1.88\%	&0.52\%\\
			\bottomrule
		\end{tabular}
	}
		
	\label{performance}
\end{table*}

\subsubsection{Backbones}
Since \themodel is plug-and-play, we plug it into several following state-of-the-art TKG reasoning models to evaluate the effectiveness of our model.

\emph{RE-NET} \cite{2020Recurrent} deals with TKGs as KG sequences. RE-NET utilizes the RGCN to capture structural dependencies of entities and relations within each KG. Then RNN is adopted to associate KGs with different time stamps for capturing the temporal dependencies of entities and relations.

\emph{RE-GCN} \cite{DBLP:conf/sigir/LiJLGGSWC21} proposes a recurrent evolution module based on relational GNNs to obtain the embeddings which contain dynamic information for entities and relations. Specially, RE-GCN designs a static module that utilizes the static properties of entities to enrich the embeddings for prediction.

\emph{CEN} \cite{DBLP:conf/acl/LiGJPL000GC22} takes the issue of length diversity in TKGs into consideration. CEN utilizes an RGCN-based encoder to learn the embeddings of entities with different history length, and adopts a CNN-based decoder to choose the optimal history length of each dataset for prediction.

\subsubsection{Evaluation Metrics}
For evaluating our model, we adopt widely used metrics \cite{2020Recurrent,DBLP:conf/sigir/LiJLGGSWC21}, MRR and Hits@\{1, 3, 10\} in experiments. To ensure the fairness of comparison among models, we unify the setting that the ground truth history is utilized during the multi-step inference for all models. Without loss of generality \cite{DBLP:conf/sigir/LiJLGGSWC21}, we only report the experimental results under the raw setting. 

\subsection{Performance Comparison (RQ1)}
Since our model utilizes multi-step updates in the testing phase, we also fine-tune all backbone models with multi-step gradient updates for a fair comparison. The performances of the backbones plugged with \themodel, the original backbones, and fine-tuned backbones on entity prediction task are shown in Table \ref{performance}. From the results in Table \ref{performance}, we have the following observations.

Our proposed method can provide significant improvements on the performance of backbones under most metrics on all datasets, which verifies the effectiveness of our method. For one thing, our method greatly enhances the performance of the original backbones, which indicates that our method can effectively help the backbones learn to adapt quickly to future data with various evolution patterns and alleviate the issue on learning entities with little historical information. For another, though fine-tuning could benefit the performance of backbones for facing new data to some degree, the backbones plugged with our model still significantly outperform the fine-tuned ones. This further illustrates the advantages of our method for learning variations in diverse evolution patterns.

Observing the results of relative improvements over the original backbones, we can find that the improvements on ICEWS05-15 are greater compared with ICEWS14, ICEWS18, and WIKI. 

It is worth noting that the number of time slices in ICEWS05-15 is much more than other datasets and facts in ICEWS05-15 last shortly \cite{han-etal-2021-learning-neural}, which implies that these facts may exhibit more diverse evolution patterns and evolve much faster in such a long time. Thus, we analyze the reason for the greater improvements on ICEWS05-15 is that the problem arising from diverse evolution patterns in ICEWS05-15 is more serious than other datasets.
Moreover, though the fine-tuned backbones can achieve great improvements on ICEWS15, our model still outperforms the fine-tuned ones. 

\subsection{Performance Comparison on Predicting Facts in Different Time Periods (RQ2)}
\begin{figure}[t]
	\centering
	\subfloat[ICEWS14]{\includegraphics[scale=0.59]{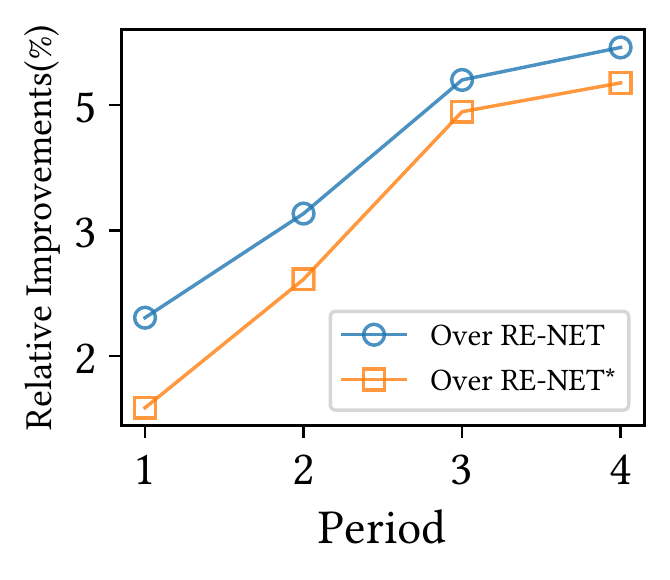}\label{period14-RENET}
	}
	\subfloat[ICEWS18]{\includegraphics[scale=0.59]{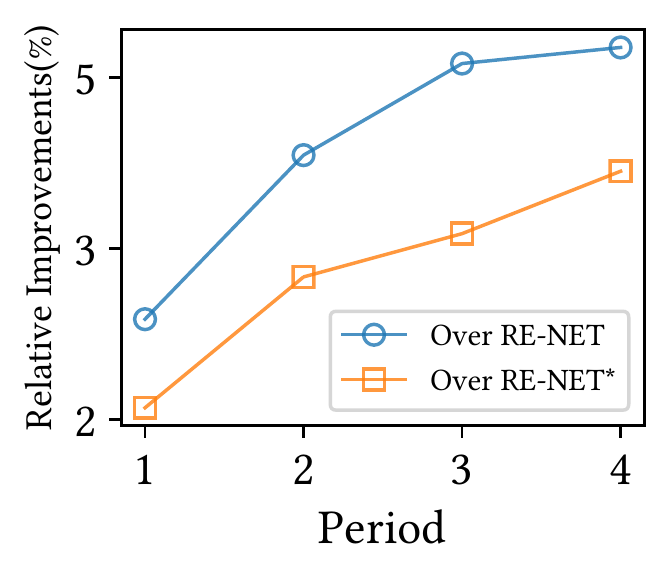}\label{period18-RENET}
	}\\
	\subfloat[ICEWS14]{\includegraphics[scale=0.59]{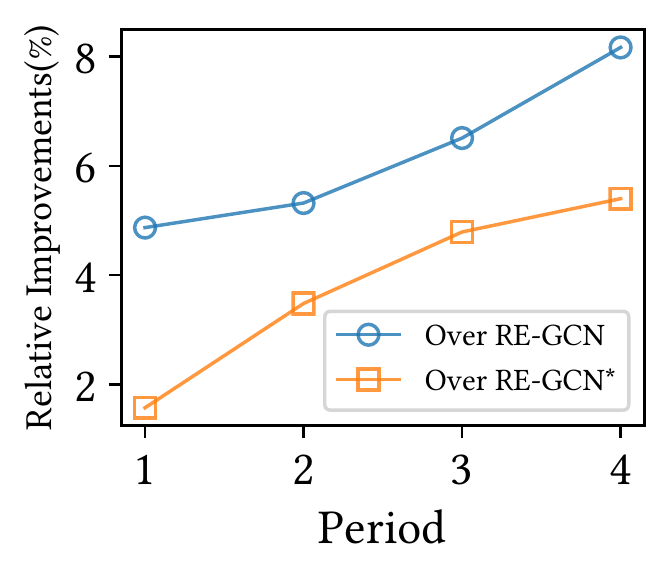}\label{period14-REGCN}
	}
	\subfloat[ICEWS18]{\includegraphics[scale=0.59]{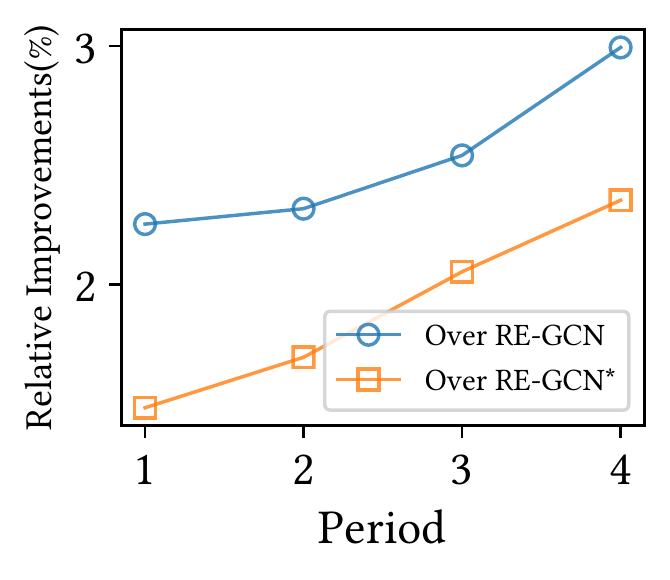}\label{period18-REGCN}
	}
	\caption{The relative improvements (MRR \%) of backbones plugged with \themodel over the original and the fine-tuned ones on predicting facts in different periods. The backbone with * represents the relative improvements (\%) over the fine-tuned one.
	}
\label{period}
\end{figure}

To verify the effectiveness of our model in solving the problem brought by diverse evolution patterns, we evaluate the performance of backbones in different periods on ICEWS14 and ICEWS18. Specifically, we divide ICEWS14 and ICEWS18 into four time periods in chronological order, respectively. In Figure \ref{period}, we represent the relative improvements of the backbone plugged with \themodel on each period to the original backbone and the fine-tuned one. From the results shown in Figure \ref{period}, we can have the following observation.

Compared with the original backbone, we can see that the relative improvement increases over time periods. Combined with Figure \ref{mov1}, we can infer that the facts in farther periods are more difficult to predict because they exhibit more different distributions than the training set. But our model can help the backbones keep effective over time because the backbones can obtain meta-knowledge from our model by training different temporal meta-tasks, which guides the backbones to learn how to adapt quickly to future data with the knowledge from old data. Moreover, compared with the fine-tuned backbone, we still can obtain the same observation described above. This illustrates that our method is much more effective in learning the variations in diverse evolution patterns, which further verifies the effectiveness of \themodel.

\subsection{Performance Comparison on Predicting Entities with Little Historical Information (RQ3)}
\begin{figure}[t]
	\centering
	\subfloat[ICEWS14]{\includegraphics[scale=0.44]{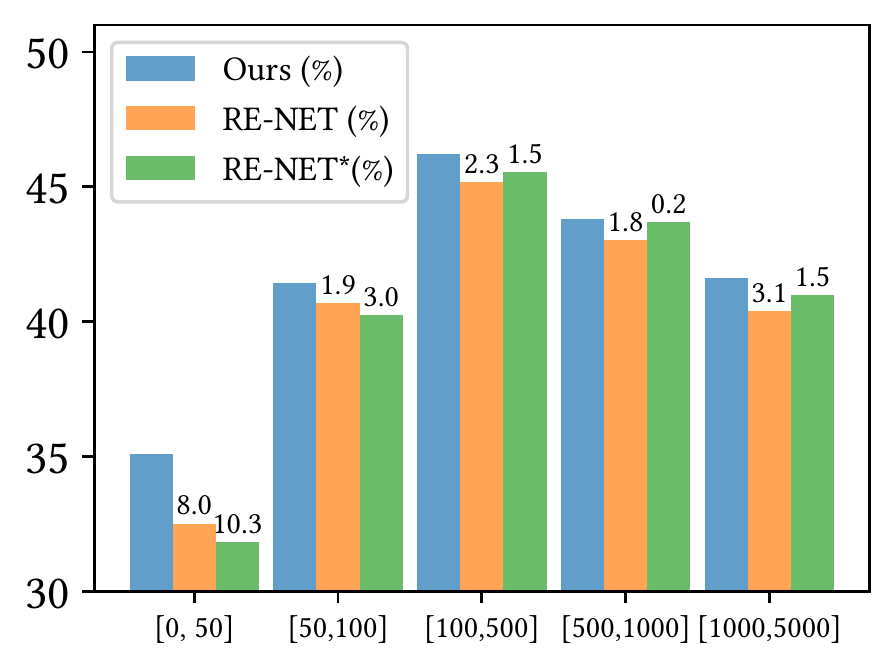}\label{few14-RENET}
	}
	\subfloat[ICEWS18]{\includegraphics[scale=0.44]{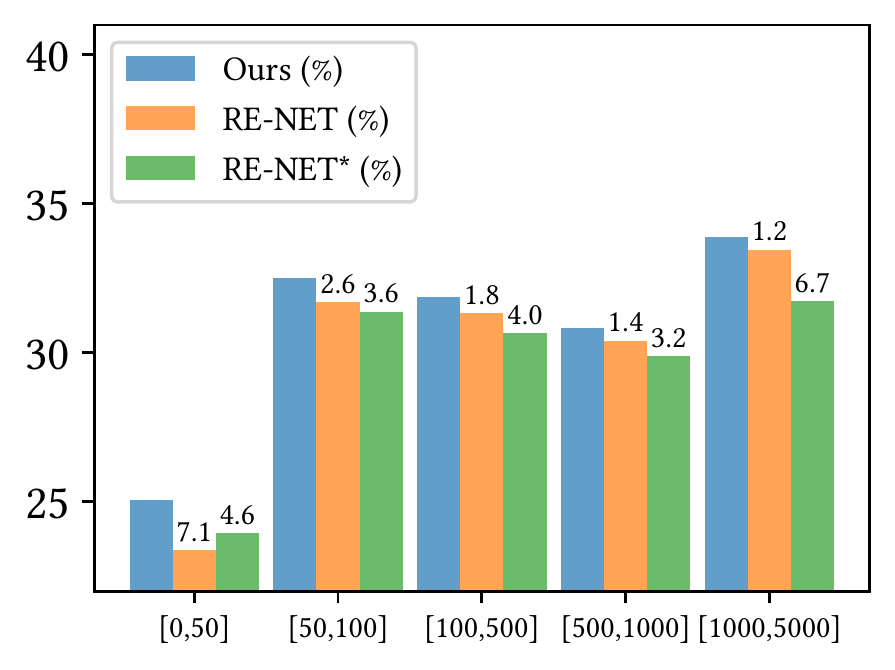}\label{few18-RENET}
	}\\
	\subfloat[ICEWS14]{\includegraphics[scale=0.44]{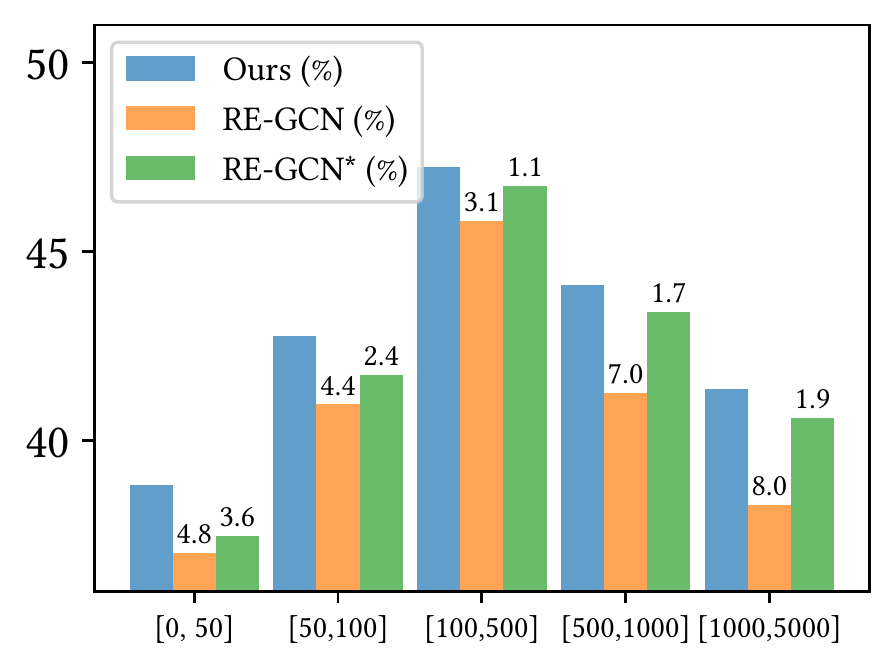}\label{few14-REGCN}
	}
	\subfloat[ICEWS18]{\includegraphics[scale=0.44]{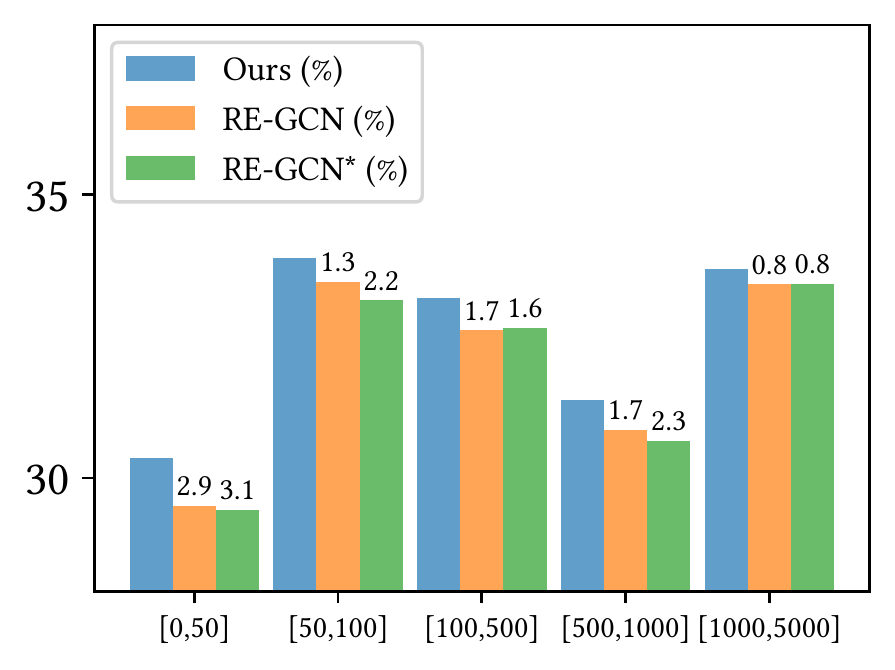}\label{few18-REGCN}
	}
	\caption{Performance comparison of models on entities with different historical interaction numbers. The y-axis is MRR value, and the x-axis represents groups with different historical interaction numbers of entities. And * represents the fine-tuned one. The numbers on each bar represent the relative improvements (\%) of the backbone plugged with \themodel to the original backbone and the fine-tuned one, respectively.
	}
\label{few-shot}
\end{figure}

We also conduct an experiment to verify the effectiveness of our model in solving the issue brought by entities with little historical information. Specifically, we divide the test sets of ICEWS14 and ICEWS18 into several groups according to the number of historical interactions of entities, respectively. Different groups contain entities with different historical interaction numbers. From the results in Figure \ref{few-shot}, we have the following observation.

Compared with both the original backbone and the fine-tuned one, the relative improvement in the group $\left[ 0, 50\right]$ is much higher than in the other groups. This indicates that both the original backbone and the fine-tuned one perform poorly on entities with little historical information, and our model is indeed effective for solving the issue of entity sparsity. We analyze the reason may be that during the training process of meta-tasks, the backbone learns to utilize the knowledge gained from entities with adequate historical information when facing entities with little historical information.

\subsection{Ablation Studies (RQ4)}
We conduct two experiments to investigate the superiority of the gating integration module in \themodel and the effectiveness of the component-specific design for the gating parameter.

\textbf{Ablation Study for Gating Module.} To verify the effectiveness of the gating integration module, we compare the performance of the backbones plugged with \themodel and \themodelAblation which removes the gating module. We show the results of all backbones plugged with the two models in Table \ref{ablation-gate} and obtain the following findings.

Firstly, all backbones plugged with \themodel outperform \themodelAblation on most evaluation metrics, which confirms that the gating integration module can effectively enhance the performance on the entity prediction task. This illustrates the importance of capturing temporal correlation in TKGs scenario by the gating module. Secondly, in the results of all datasets for each backbone, the relative improvement on WIKI is lower compared with other datasets. For WIKI, it is worth noting that facts in WIKI last much longer and do not occur periodically \cite{han-etal-2021-learning-neural}, which implies that these facts may change slowly. Thus, this observation may indicate that the gating module is more capable of establishing temporal correlations for datasets in which facts evolve significantly over time.   

\textbf{Ablation Study for Component-specific gating parameter.} The gating parameter is the key to determining the performance of the gating integration module. In this experiment, we compare the performance of the backbones plugged with \themodel and MetaTKG-C, in which the gating parameter $\operatorname{g}_s$ is set to the same value for entity embedding, relation embedding, and other parameters in the model. We show the results of all backbones plugged with the two models in Table \ref{ablation-g}.

Through the relative improvements, we can find that assigning different $\operatorname{g}_s$ values to entity embedding, relation embedding and other parameters in the model can achieve more superior performance on all datasets with different backbones. This demonstrates the robustness and effectiveness of our $\operatorname{g}_s$ design.

\begin{table}[t]
	\centering
	\caption{Ablation studies on gating integration module in terms of MRR (\%) under the raw setting.}
	\resizebox{\linewidth}{!}
	{\begin{tabular}{ccccc}
			\toprule
			 Model &  ICEWS14 &  ICEWS05-15 & ICEWS18 &  WIKI \\
			\midrule
			 RE-NET + MetaTKG &\textbf{40.87}	&\textbf{30.19}	&\textbf{49.15}	&\textbf{53.28} \\
			 RE-NET+MetaTKG-G &39.72 &29.18	&48.01	&52.54\\
			 \rowcolor{gray!20}$\Delta$\emph{Improve} &2.90\%	&3.46\%	&2.37\%	&1.41\% \\
			\midrule
			RE-GCN + MetaTKG  &\textbf{42.76}	&\textbf{31.60}	&\textbf{50.00}	&\textbf{53.54}\\
			RE-GCN + MetaTKG-G &41.54	&31.14	&49.36	&53.05\\
			\rowcolor{gray!20}$\Delta$\emph{Improve} &2.94\%	&1.48\%	&1.30\%	&0.92\%\\
			\midrule
			CEN + MetaTKG &\textbf{41.64}	&\textbf{29.70}	&\textbf{48.77}	&\textbf{53.42}\\
			CEN + MetaTKG-G &40.86	&28.98	&48.21	&52.98\\
			\rowcolor{gray!20}$\Delta$\emph{Improve} &1.91\%	&2.48\%	&1.16\%	&0.83\%\\
			\bottomrule
	\end{tabular}}
	\label{ablation-gate}
\end{table}

\begin{table}[t]
	\centering
	\caption{Ablation studies for component-specific gating parameter in terms of MRR (\%) under the raw setting.}
	\resizebox{\linewidth}{!}
	{\begin{tabular}{ccccc}
			\toprule
			 Model &  ICEWS14 &  ICEWS05-15 & ICEWS18 &  WIKI \\
			\midrule
			 RE-NET + MetaTKG &\textbf{40.87}	&\textbf{30.19}	&\textbf{49.15}	&\textbf{53.28} \\
			 RE-NET+MetaTKG-C &40.34 &29.98	&48.51	&52.78\\
			 \rowcolor{gray!20}$\Delta$\emph{Improve} &1.31\%	&0.70\%	&1.32\%	&0.95\% \\
			\midrule
			RE-GCN + MetaTKG  &\textbf{42.76}	&\textbf{31.60}	&\textbf{50.00}	&\textbf{53.54}\\
			RE-GCN + MetaTKG-C &41.98	&31.31	&49.52	&53.21\\
			\rowcolor{gray!20}$\Delta$\emph{Improve} &1.86\%	&0.93\%	&0.97\%	&0.62\%\\
			\midrule
			CEN + MetaTKG &\textbf{41.64}	&\textbf{29.70}	&\textbf{48.77}	&\textbf{53.42}\\
			CEN + MetaTKG-C &40.98	&29.21	&48.39	&53.02\\
			\rowcolor{gray!20}$\Delta$\emph{Improve} &1.61\%	&1.68\%	&0.79\%	&0.75\%\\
			\bottomrule
	\end{tabular}}
	\label{ablation-g}
\end{table}

\subsection{Analysis on the effect of the number of gradient update steps (RQ5)}
\begin{figure}[t]
	\centering
	\subfloat[ICEWS14]{\includegraphics[scale=0.62]{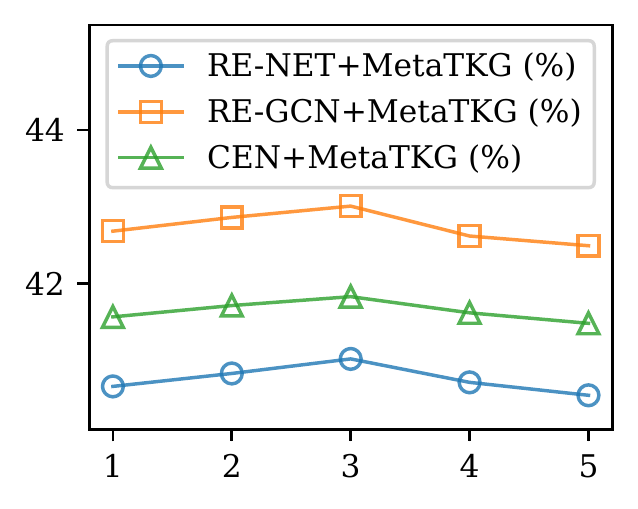}\label{multi14}
	}
	\subfloat[ICEWS18]{\includegraphics[scale=0.62]{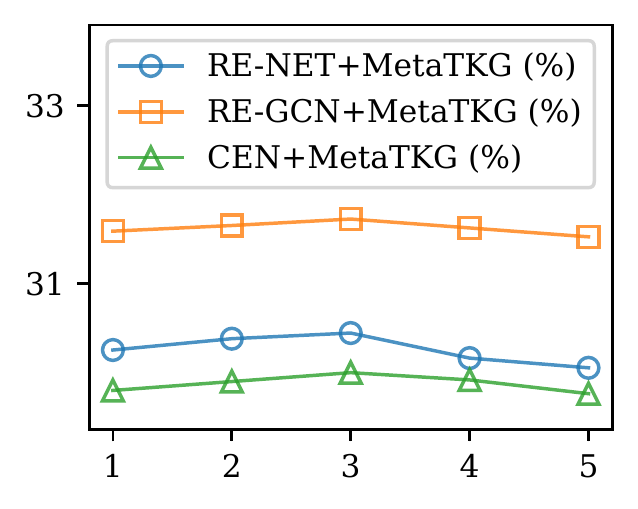}\label{multi18}
	}\\
	\subfloat[ICEWS05-15]{\includegraphics[scale=0.62]{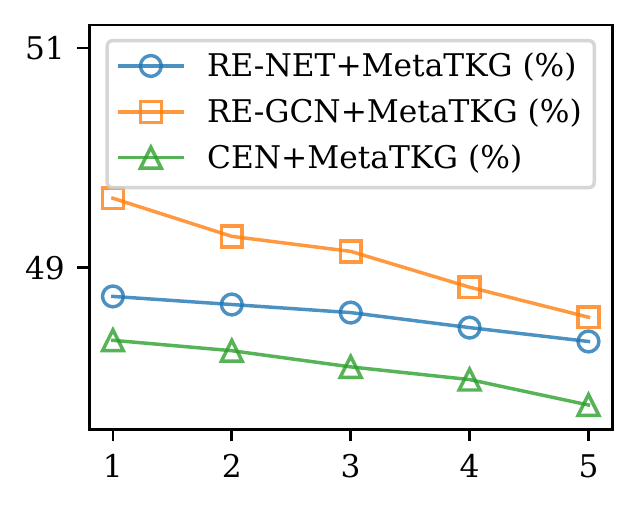}\label{multi15}
	}
	\subfloat[WIKI]{\includegraphics[scale=0.62]{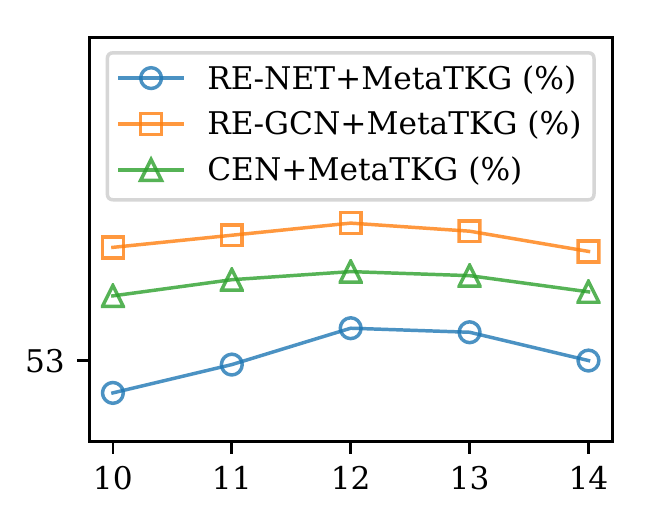}\label{multiWIKI}
	}
	\caption{Effect of different gradient update step numbers in the testing phase. The y-axis is MRR value. And the x-axis represents the different number of gradient update steps.
	}
\label{multi}
\end{figure}

In this section, we study how the multi-step gradient updates affect our model. The performances of backbones plugged with \themodel under different multi-step values are shown in Figure \ref{multi}. We can find that the multi-step updates can enhance the performance of backbones on most datasets. However, compared with the other three datasets, the multi-step updates seem to be less effective on ICEWS05-15. From Table \ref{tab:dataset}, we can find that the number of time slices in ICEWS05-15 is much more than other datasets, but the total number of facts is not larger than others, which indicates that the number of facts in ICEWS05-15 in each time slice is relatively less. Thus, we analyze the reason for the aforementioned observation is that the multi-step updates make the models susceptible to over-fitting when predicting each time slice.

\section{Conclusion}
In this paper, we have proposed a novel temporal meta-learning framework \themodel for TKG reasoning, which can be easily served as a plug-and-play module for most existing TKG prediction models. \themodel regards TKG prediction as temporal meta-tasks, and utilizes a Temporal Meta-Learner to learn the evolutionary meta-knowledge from these tasks. \themodel aims to guide the backbones to adapt quickly to future data and enhance the performance on entities with little historical information by the learned meta-knowledge. Specially, in the temporal meta-learner, we develop a Gating Integration module to establish temporal correlations between tasks. Extensive experiments on four benchmarks and three state-of-the-art backbones for TKG prediction demonstrate the effectiveness and superiority of \themodel.

\section*{Limitations}
In this section, we discuss the limitation of our model. Specifically, we utilize multi-step gradient updates in the testing phase to enhance the ability of fast adaption for the backbones plugged with our model. In this way, the effectiveness of our model can be more significantly improved compared with one gradient update, but the requirement of GPU resources becomes larger than the original backbones. Moreover, since the model acts on data in one time slice at a time for TKG prediction, multi-step gradient updates tend to cause over-fitting on the dataset in which the number of data in each time slice is small, such as ICEWS05-15.

\section*{Acknowledgements}
This work is sponsored by National Natural Science Foundation of China (NSFC) (U19B2038, 62141608, 62206291, 61871378, U2003111), and Defense Industrial Technology Development Program (Grant JCKY2021906A001).

\bibliography{ref}
\bibliographystyle{acl_natbib}

\section*{Appendix}
\appendix

\section{Discussion on Temporal Correlation}
\label{temproal connetion}
The meta-task in traditional static scenes is usually independent of each other, especially there are no temporal associations among them in static scenes. Different from this, there are temporal connections among tasks in the scenario of TKGs. For example, the task $\mathcal{T}_1$ is composed of $\mathcal{G}_1$ and $\mathcal{G}_2$, and $\mathcal{T}_2$ consists of $\mathcal{G}_2$ and $\mathcal{G}_3$. Obviously, there are temporal connections among time slices ${G}_1$, ${G}_2$ and ${G}_3$ due to the time-series characteristics of TKG, so the temporal meta-tasks composed of adjacent time slices are certainly temporally associated. 

Considering the temporal connection among tasks, the training for the next task may need the knowledge learned from former ones. The easiest way to achieve this is training tasks one by one and initializing the parameter of $\mathcal{T}_t= \{\mathcal{G}_t, \mathcal{G}_{t+1}\}$ with the updated parameter by $\mathcal{T}_{t-1} = \{\mathcal{G}_{t-1}, \mathcal{G}_{t}\}$. However, temporal connections established in this way are not effective enough because the temporal information between adjacent time slices is unable to utilize. Note that the initial parameter is operated on the support set and the final parameter updated by $\mathcal{T}_t$ is actually operated on the query set of $\mathcal{T}_t$. That is, if we build up connection in this way, the support set $\mathcal{G}_t$ of the task $\mathcal{T}_t$ will be temporally associated with $\mathcal{G}_t$ instead of $\mathcal{G}_{t-1}$.

\section{Implementation Details}
\label{implementation details}
We implement our \themodel in \emph{Pytorch} \cite{Paszke:2019vf}. We optimize all models with Adam optimizer \cite{Kingma:2015us}, where the learning rates of both support and query set are $0.001$ and $l_2$ regularization $\lambda_2$ is set to $10^{-5}$. We use the Xavier initializer \cite{DBLP:journals/jmlr/GlorotB10} to initialize parameters. When training, we only conduct a one-step gradient in temporal meta-learner for efficiency. In the testing phase, we conduct multi-step gradient update. We perform grid search for the multi-step on validation set. The optimal multi-step for ICEWS14, ICEWS18, ICEWS05-15, and WIKI are 3, 3, 1, and 12, respectively. The optimal values of multi-step are set the same on all backbone models. All experiments are conducted on NVIDIA Tesla V100 (32G) and Intel Xeon E5-2660.

\end{document}